\documentclass{article}
\usepackage[a4paper, margin=1in]{geometry}
\usepackage{cite}
\usepackage{amsmath,amssymb,amsfonts}
\usepackage{algorithmic}
\usepackage{graphicx}
\usepackage{textcomp}
\usepackage{booktabs}
\usepackage{tabularx}
\usepackage{multirow}
\usepackage{svg}
\usepackage{hyperref}
\usepackage{threeparttable}
\usepackage{hyperref}

\begin{document}
\title{Masked Pre-Training of Transformers for Histology Image Analysis}
\author{Shuai Jiang\thanks{Shuai Jiang, Liesbeth Hondelink, and Saeed Hassanpour are with the Department of Biomedical Data Science, Geisel School of Medicine at Dartmouth, Hanover, NH 03755 USA (e-mail: Shuai.Jiang.GR@dartmouth.edu, l.m.hondelink@lumc.nl, Saeed.Hassanpour@dartmouth.edu).}, Liesbeth Hondelink\footnotemark[1], Arief A. Suriawinata\thanks{Arief A. Suriawinata is with the Department of Pathology and Laboratory Medicine, Dartmouth-Hitchcock Medical Center, Lebanon, NH 03756, USA (e-mail: arief.a.suriawinata@hitchcock.org).}, Saeed Hassanpour\footnotemark[1] \thanks{Saeed Hassanpour is also with the Department of Epidemiology of Geisel School of Medicine at Dartmouth and the Department of Computer Science, Dartmouth College, Hanover, NH 03755, USA.}
\thanks{This research was supported in part by grants from the US National Library of Medicine (R01LM012837 and R01LM013833) and the US National Cancer Institute (R01CA249758).\textit{(Corresponding author: Saeed Hassanpour)}}
}
\date{}
\maketitle

\begin{abstract}
In digital pathology, whole slide images (WSIs) are widely used for applications such as cancer diagnosis and prognosis prediction. Visual transformer models have recently emerged as a promising method for encoding large regions of WSIs while preserving spatial relationships among patches. However, due to the large number of model parameters and limited labeled data, applying transformer models to WSIs remains challenging. Inspired by masked language models, we propose a pretext task for training the transformer model without labeled data to address this problem. Our model, MaskHIT, uses the transformer output to reconstruct masked patches and learn representative histological features based on their positions and visual features. The experimental results demonstrate that MaskHIT surpasses various multiple instance learning approaches by 3\% and 2\% on survival prediction and cancer subtype classification tasks, respectively. Furthermore, MaskHIT also outperforms two of the most recent state-of-the-art transformer-based methods. Finally, a comparison between the attention maps generated by the MaskHIT model with pathologist's annotations indicates that the model can accurately identify clinically relevant histological structures in each task.
\end{abstract}


\section{Introduction}
\label{sec:introduction}
Histopathological slides are widely used in clinical settings for cancer diagnosis, prognosis, and treatment planning. These slides reveal crucial information such as the presence, type, degree of differentiation, and mitotic activity of tumor cells, as well as the size, growth, and necrosis of tumor tissue when examined under a high-magnification microscope. Histopathological examination is widely regarded as the gold standard for cancer diagnosis and subtype determination. Although manually curated histopathological features from pathologists' inspection can improve prognosis prediction accuracy \cite{Romansik2007,Azzola2003}, the process is time-consuming, subjective \cite{Rousselet2005, Veta2016}, and highly specialized, hindering the widespread use of histopathological features in clinical applications, particularly in developing countries or rural settings \cite{benediktsson2007pathology}. Whole slide images (WSIs) are digitized versions of conventional slides, making it easier to store and share histopathological information. The increasing popularity of WSIs also facilitates the development of computational pathology solutions that focus on the automatic interpretation of histopathological images using computational approaches, such as deep learning \cite{Abels2019,Nasir-Moin2021}. One group of computational pathology tasks is segmenting tumor regions on a WSI \cite{Xu2017,Qaiser2019,Wei2019,Wei2020}. Another group of tasks involves providing whole slide-level inferences, such as providing diagnosis and prognosis information \cite{Tomita2022,Yao2020,Lu2021,DiPalma2021,Kim2022,Barrios2022,Zhu2021}. This paper will focus on the second group of tasks.

As a result of the remarkable success of Convolutional Neural Network-based (CNN) models, such as Residual Network (ResNet), in the image analysis realm \cite{He2016}, a large number of studies have adopted these models for automatic feature extraction from WSIs to inform patient diagnosis and prognosis \cite{Wulczyn2020,Dimitriou2019}. However, feeding the entire WSI into a machine learning model is not feasible due to WSI size and current hardware capacities \cite{Wang2012,araujo2019computing}. A typical WSI can be extremely large, containing billions of pixels, which is several orders of magnitude larger compared to natural images, making it impossible to build a CNN with an appropriate receptive field while maintaining manageable GPU memory usage. Alternatively, a multi-step method is usually adopted that analyzes a WSI in smaller portions instead of all at once \cite{Wulczyn2020,Veta2019}. The first step in these pipelines normally involves dividing the WSIs into smaller units known as patches, typically with a size of several hundred pixels (e.g., 224$\times$224). A pre-trained feature extractor is then employed to transform these patches into feature vectors. The next step involves utilizing an aggregation method to derive the overall representation of the WSI from the feature vectors of several sampled patches. Finally, this global representation vector is used to make predictions at the slide level.

The unique challenge posed by WSIs is the need to integrate high-level features from numerous smaller images. Because labels are typically only available at the slide or patient level, this aggregation is considered a form of weakly supervised learning. The most widely employed approach in this field is Multiple Instance Learning (MIL), which treats a WSI as a bag of patches and neglects the positional information and relationships among those patches. Extensions of the MIL approach utilize the attention mechanism, which assigns weights to patches in the aggregation process based on their corresponding visual patterns \cite{Yao2020,Ilse2018}. One significant drawback of these techniques is that they do not consider the spatial arrangements and associations of the patches and their patterns, which can limit the recognition and integration of features beyond a single patch \cite{Dimitriou2019}.

In this study we developed a method named Masked Pre-training for Histology Images using Transformer (MaskHIT), as a pre-training and fine-tuning pipeline for whole-slide level representation and analysis. Our proposed methodology can effectively integrate both low- and high-level features from a WSI with a hybrid approach, with the low-level features extracted using pre-trained CNN models from tissue patches, while the high-level features represented with vision transformer model that facilitates learning the relationships among patches. The self-attention mechanism and positional encoding capabilities of the transformer model allow for the integration of information spanning beyond a single patch. To enhance the performance of the transformer model, we undertake a pre-training step using a subset of The Cancer Genome Atlas (TCGA) database through the masked patch prediction method. This inpainting procedure involves randomly obscuring a portion of the patches and training the model to reconstruct the masked patches. While fine-tuning for downstream tasks, the predictions generated by the model can also be explained by highlighting the regions of the WSI that have the most significant impact on the prediction. These indicative patches can be visualized and reviewed by clinicians to facilitate the recognition of the associations between various morphological attributes and the slide-level labels in research and clinical practice. We have made the code of this study publicly available at \url{https://github.com/BMIRDS/WSI-PLP/tree/maskhit}.

\section{Related Work}
\subsection{Multiple Instance Learning for WSIs}

One fundamental step in WSI-based prediction tasks is aggregating patch-level information to make slide-level predictions. Such aggregation can be performed either with patch-level features or patch-level predictions. For example, the maximum value from patch predictions was used to make slide-level classification predictions, assuming the presence of at least one positive instance when the slide is positive \cite{Campanella2019}. On the other hand, for prognosis prediction tasks, average pooling over patch features is commonly used \cite{Wulczyn2020,Jiang2021}. However, these fixed operations lack the required adaptability for different domains.

More sophisticated approaches have been developed to enable feature aggregation from multiple patches with a learnable operation. For example, a 3D convolutional filter was used to compute the attention score of each patch for patch feature aggregation \cite{Tomita2019}. In the MIL field, Ilse et al. implemented a flexible module that assigns learnable weights to each instance and aggregates patch features dynamically through a weighted average approach \cite{Ilse2018}. This attention-based method was later expanded upon by \cite{Lu2021}, in which a clustering loss was employed to impose constraints on the feature space identified by the attention module for slide-level classification tasks.

The attention-based approach was later adapted for predicting survival outcomes as well. Yao et al. utilized a multi-cluster approach, where each patch was categorized into one of several clusters and an attention layer was employed to integrate features from each cluster \cite{Yao2020}. Previously, our team introduced a multiple-head attention approach, which enables multiple attention layers to work in collaboration to achieve a comprehensive feature extraction \cite{Jiang2023}. Despite these advancements, it must be noted that these multiple instance-based methods do not account for the spatial configuration of patches, which can limit the effectiveness of the models.

\subsection{Vision Transformer in Computational Pathology}

The transformer model was introduced in the realm of language modeling and specifically utilized for sequence-to-sequence learning tasks such as machine translation \cite{Vaswani2017}. One of the defining features of the transformer model is the incorporation of the self-attention mechanism, which allows each element within a sequence, such as a word within a sentence, to consider information from all other elements in the sequence and determine their contributions to the encoding of the current element. Additionally, positional encoding is utilized to capture the relative position of each element within the sequence. Given the vast number of parameters involved in this model, a pretext task was devised to train the model in an unsupervised manner. The BERT language model was developed to leverage the transformer architecture and utilized masked word prediction and next sentence prediction tasks for training purposes. This approach yielded remarkable results and new advances in natural language processing \cite{Devlin2019}. 

Moreover, the transformer framework has also been adapted for computer vision tasks. The recently introduced Vision Transformer (ViT) model has demonstrated exceptional performance in image classification tasks without relying on CNNs \cite{Dosovitskiy2020}. Transformer model has been introduced to the digital pathology field as well. For example, multiple studies have used the transformer model as an aggregation method under the framework of MIL for whole slide-level prediction tasks \cite{Li2023,Chen2022}. Transformer model has also been pre-trained using a contrastive learning approach to replace ResNet as the backend for patch level feature extraction \cite{Mistry2003}. In addition, Chen et al. implemented hierarchical transformer modules (HIPT) that can be pre-trained to learn high-resolution representations of WSIs \cite{Chen2022hipt}. HIPT pre-training adopted the DINO framework with a student and a teacher network, and the model is trained so that the probability distribution of the output from the student network can match that from the teacher network. 

\begin{figure*}[t]
    \centering
    \includegraphics[width = \textwidth]{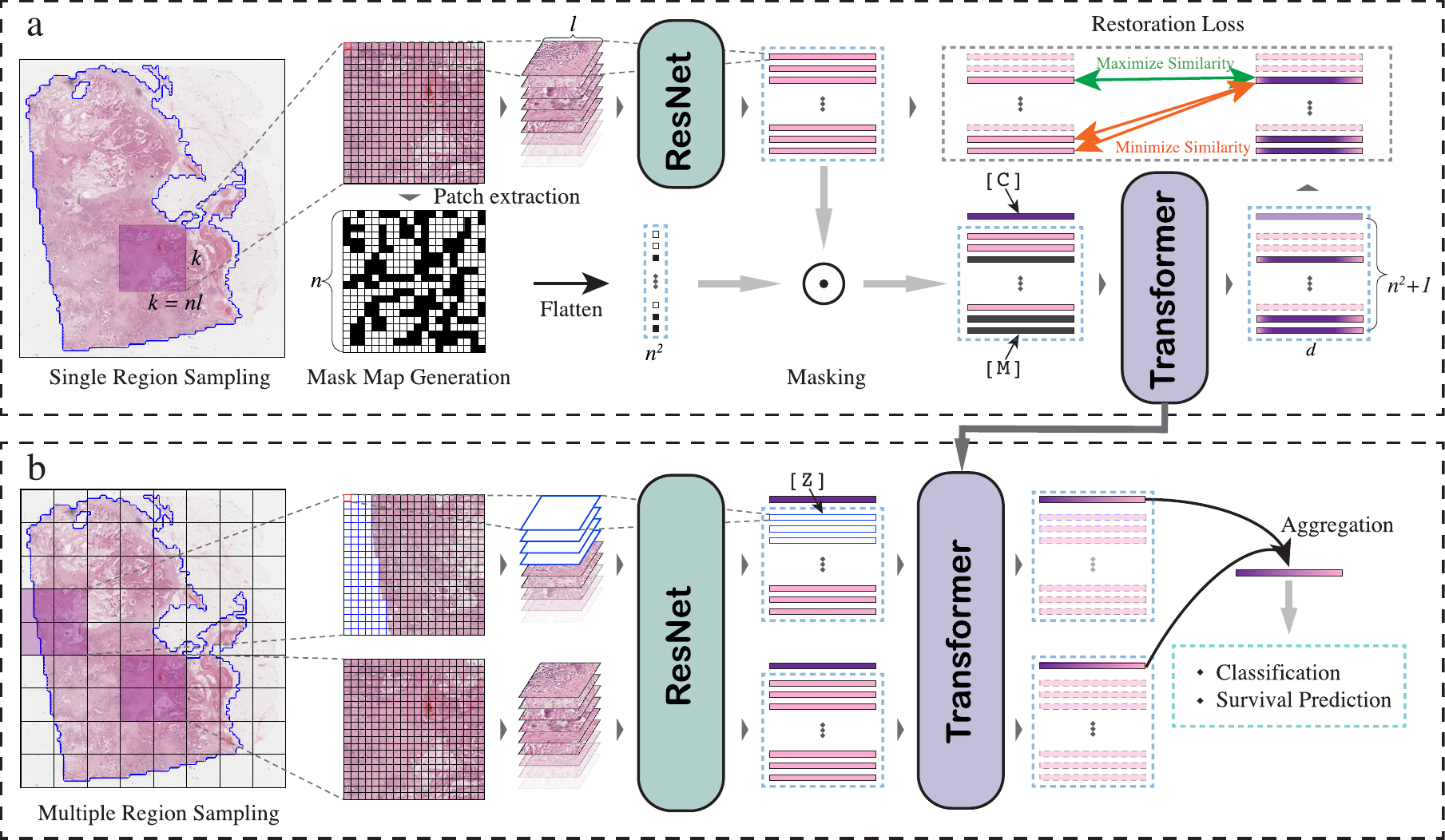}
    \caption{MaskHIT Architecture. a) Pretraining stage. b) Fine-tuning stage. [M]: mask token. [C]: class token. [Z]: zero padding.}
\end{figure*}

\section{Methods}

An overview of our pipeline is shown in Figure 1. We first randomly choose a region from WSI and break it down into non-overlapping patches. A pre-trained ResNet model serves as a fixed feature extractor for patch processing, converting each patch into a feature vector. Meanwhile, a random mask is generated for the region and is used to mask out the preprocessed feature matrix. This masked input is then sent to the transformer model along with positional information. The output from the transformer model is used to restore the original unmasked patch features. Following the pre-training step, regions are sampled from the WSI randomly or through systematic sampling procedures, and no masking is performed at this stage. Those rgions are then processed via the pre-trained transformer model, and their output class tokens are aggregated for downstream tasks.

\subsection{Hybrid representation of whole slide image}
In accordance with MIL framework for WSIs, a WSI, $\mathcal{X}$, can be viewed as a bag of patches:
\begin{equation}\mathcal{X}=\{\mathcal{X}_1^l,\mathcal{X}_2^l,...,\mathcal{X}_N^l\}\label{eq}\end{equation}
where $\mathcal{X}_i^l$ refers to one patch with size $l\times l$ (e.g., 224$\times$224) pixels selected from the WSI and $N$ is the total number of patches from this WSI. However, as this approach treats the patches as independent, their spatial relationships are ignored during modeling. We sought to address this problem by representing the whole slide as $M$ large regions of size $nl\times nl$ (e.g., 4480$\times$4480 when $n$=20 and $l$=224) pixels that contain high-level structure of a tissue.
\begin{equation}\mathcal{X}=\{\mathcal{X}_1^k,\mathcal{X}_2^k,...,\mathcal{X}_M^k\}\label{eq}\end{equation}
and
\begin{equation}
\mathcal{X}_i^k=\mathcal{G}((\mathcal{X}_{i,1}^l,\mathcal{P}_{i,1}),(\mathcal{X}_{i,2}^l,\mathcal{P}_{i,2}),…,(\mathcal{X}_{i,n^2}^l,\mathcal{P}_{i,n^2}))
\label{eq}\end{equation}
where $\mathcal{P}_{i,j}\in \mathbb{R}^2$ is the position of the $j$th patch on region $i$. We choose $\mathcal{G}$ to be the vision transformer model as it can effectively model the positions of patches.

In a typical vision transformer model, an image with a regular size is first split into small patches \cite{Dosovitskiy2020}. The patches are flattened into a column vector, positionally encoded, and then fed to a multiple-layer transformer model together with a special token (\textit{class} token) to capture the global information of the image. This process naturally fits the multi-stage framework of modeling WSIs. Similarly, we can sample a large ``image" $\mathcal{X}^k$ from WSI, then apply the pre-trained ResNet model for patch feature extraction and the transformer model for patch feature aggregration. This hybrid approach combining a CNN model with a transformer encoder has been shown to deliver improved performance for datasets with small-to-moderate sample sizes \cite{Dosovitskiy2020}, and can also help us focus on the training of the transformer part of the pipeline.

For each patch, we first add the feature vector with a positional vector \cite{Vaswani2017, Gehring2017}. For the feature representation of the patches with a dimension of d, we randomly initialize two learnable positional matrices of size $\mathcal{P}_x,\mathcal{P}_y \in \mathbb{R}^{n\times(d/2)}$ for x- and y-axis locations. At position $\mathcal{P}_{i,j}=(x_j,y_j)$, the positional representation of the patch is the concatenation of x- and y-axis positional encodings: $concat(\mathcal{P}_x [x_j],\mathcal{P}_y [y_j])$. A learnable class token is appended to the start of transformer's input sequence. This token serves as the high-level representation of a WSI during training and can be used as the region-level feature vector for downstream tasks.

This sequence is inputted to an $L$-layer transformer encoder with each layer containing an $H$-head self-attention module \cite{Vaswani2017}. Patches that are not from the foreground regions are zero-padded. We apply masked multi-head self-attention in the transformer model to assign a pre-normalized activation value of $-1\times10^5$ for the locations that are masked as background, so their corresponding attention weight will essentially be 0 and the attention weights for the tissue patches will be correctly normalized. 

\subsection{Pre-training with Masked Patch Prediction}
Given the large number of parameters in the multi-layer transformer model, and the relatively small size of labeled WSI datasets, training the transformer model directly for outcome prediction would lead to serious overfitting problems during optimizing a transformer model. Therefore, we first pre-train the transformer model in an unsupervised manner before optimizing it for downstream tasks. This approach can greatly benefit from publicly available, unlabeled datasets. Our masked patch prediction method for pre-training operates as follows.

First, we apply a random masking procedure on patches with a fixed probability of $p$ utilizing the blockwise masking technique \cite{Bao2021}. The technique involves masking one to four patches at a time until the desired masking rate of $p$ is achieved. The feature vectors of the masked patches are then substituted with a learnable mask token. The positional encoded patch feature sequence and masked patches are fed into the transformer model. The restoration loss is calculated based on the output from the last layer of the transformer model at the masked locations. The unmasked patches and background patches are excluded from the pretext loss calculation. All selected patches from a batch are mixed in a single list of length $k$, regardless of their region or slide membership. Here, $y_i\in \mathbb{R}^d$ refers to the $i$th output vector from the transformer model, and $x_i \in \mathbb{R}^d$ refers to the corresponding unmasked feature vector of this patch.

Our loss function consists of two essential components. The first component aims to maximize the similarity between the output features of the masked patches and their original ResNet representations. This is achieved by calculating the $L2$ distance between the output from the final layer and the original unmasked feature representation. The second component is a modified version of the InfoNCE loss, which can minimize the similarity between an output instance and all other selected input feature sequences \cite{Oord2018,Le-Khac}. For $i$th entry in the mixed output, the loss is expressed as
\begin{equation}\mathcal{L}_i=\alpha\lVert x_i - y_i\rVert+\beta l(x_i,y_i)\label{eq}\end{equation}
where
\begin{equation}
l(x_i,y_i)=-\log\frac{\exp(\lVert x_i - y_i\rVert/\tau)}{\Sigma_{j=1}^k{\exp(\lVert x_j - y_i\rVert/\tau)}}
\label{eq}\end{equation}
and $\alpha$ and $\beta$ values are determined through a grid search method (Section \ref{sssec:alpha-beta}). The temperature parameter $\tau$ is set to be 0.1 following a previous study \cite{Caron2020}.
\subsection{Fine-tuning for whole-slide level predictions}
After pre-training the transformer model using large unlabeled datasets, we can fine-tune the model for the specific tasks of interest with datasets not used in the pre-training. To accomplish this, we select a fixed number of regions from the WSI and feed them to the pre-trained transformer model without undergoing patch masking. Upon obtaining the class tokens from each region, a straightforward aggregation method is employed to calculate an average of these class tokens, which serves as the representation vector for the entire slide. This average class token can then be utilized to make slide-level predictions. A lower learning rate is assigned to the transformer component of the model to allow gradual gradient-based updates for downstream tasks.

\section{Experiments}

We have conducted two series of experiments to evaluate MaskHIT. In the first series of experiments, we pre-trained the MaskHIT model with ResNet-18 features at 10$\times$ magnification (MaskHIT-10$\times$) using 5 cancer datasets from TCGA, and then compared it on downstream cancer survival prediction (\textit{Survival-1}) and classification (\textit{Classification-1}) tasks to other MIL-based approaches. We also conducted ablation experiments to verify the model design choices. In the second series of experiments, we pre-trained the MaskHIT model with ResNet-34 features at 20$\times$ magnification (MaskHIT-20$\times$). 14 cancer datasets from TCGA are used in this pre-training. Then we compared our model to the latest transformer-based approaches on a different set of tasks (\textit{Survival-2} and \textit{Classification-2}).
\subsection{Datasets}

\autoref{table:data} summarizes the information about the datasets that we used in our experiments. Our experiments were limited to utilizing only Formalin-Fixed Paraffin-Embedded (FFPE) slides to avoid the excess artifacts on frozen slides.

Pre-training of the MaskHIT-10$\times$ model utilized 5 cancer datasets. These datasets consist of 2244 slides and about 20 million patches. Pre-training of the MaskHIT-20$\times$ model was extended to 14 cancer types from TCGA which were about 2 times more data than MaskHIT-10$\times$ model, as indicated in \autoref{table:data}.

Survival-1 tasks utilized 5 cancer datasets from TCGA that were not included in training the MaskHIT-10x model, which are breast cancer (BRCA), colon adenocarcinoma (COAD), brain lower grade glioma (LGG), lung adenocarcinoma (LUAD), and ovarian serous cystadenocarcinoma (OV). The three tasks of Classification-1 are: 1) glioblastoma multiforme (GBM) versus LGG classification; 2) BRCA molecular subtypes classification, namely, HER2-enriched, luminal A, luminal B, and basal-like; and 3) Renal Cell Carcinoma (RCC) subtype classification, which are clear cell, papillary, chromophobe, benign, and oncocytoma. The final task used one internal dataset from Dartmouth Hitchcock Medical Center (DHMC) \cite{Zhu2021}, whereas all other tasks used TCGA datasets.

Survival-2 and Classification-2 tasks followed the setup in HIPT \cite{Chen2022hipt} for equitable comparison. There are 6 survival tasks, namely clear cell renal cell carcinoma (CCRCC), stomach adenocarcinoma (STAD), invasive ductal carcinoma (IDC), colorectal cancer (CRC), LUAD, and papillary renal cell carcinoma (PRCC). Moreover, there are 3 classification tasks, namely, Breast (invasive ductal versus invasive lobular), Lung (adenocarcinoma versus squamous cell carcinoma), and Kidney (clear cell versus papillary versus chromophobe).

The tissue mask of WSIs was calculated at a low magnification level (32 \textmu m per pixel, or 0.3125$\times$) using the purple thresholding method for thumbnail images and further processed with binary dilation and erosion to eliminate small holes and debris. We then sequentially extracted patches at 10$\times$ (1 \textmu m per pixel) and 20$\times$ (0.5 \textmu m per pixel) from WSIs without overlap. At least 5\% of pixels within a patch had to be from tissue for that patch to be considered foreground. ResNet-18 (or ResNet-34 for 20$\times$) model pre-trained using the ImageNet was used as the fixed feature extractor to process all the foreground patches. The extracted feature vectors, with dimension of 512, were saved on storage before the experiments to cut on computational cost during model training and testing. Background patches were excluded from the feature extraction step and padded with zero vectors. Also, the position of each patch within the region $(x, y)$ was recorded.

\begin{table}[]
\centering
\caption{Summary of datasets and usage in experiments.}
\label{table:data}

\begin{tabular}{llccccp{15mm}}
\toprule 
& Experiments          & Tasks & Datasets & Slides & Regions & Patches \newline ($\times10^6$) \\
\midrule
\multirow{2}{*}{PT} & MaskHIT-10$\times$          & 1     & 5        & 2,244   & 35,509   & 19.810         \\
                              & MaskHIT-20$\times$       & 1     & 14       & 7,095   & 99,134   & 60.686         \\
                              \midrule
\multirow{4}{*}{FT}   & Survival-1       & 5     & 5        & 3,065   & 39,408   & 26.880         \\
                              & Classification-1 & 3     & 4        & 3,308   & 41,140   & 26.481         \\
                              & Survival-2       & 6     & 7        & 3,513   & 47,091   & 31.344         \\
                              & Classification-2 & 3     & 6        & 3,108   & 42,003   & 25.781         \\ 
                              \bottomrule
\multicolumn{7}{l}{
  \begin{minipage}{11cm}%
  \bigskip
    \footnotesize
    \hspace{-2.5pt}\noindent At 10$\times$ region size is 4480$\times$4480 pixels and patch size is 224$\times$224 pixels. At 20$\times$ region size is 8960$\times$8960 pixels and patch size is 448$\times$448 pixels.
  \end{minipage}%
}
\end{tabular}
\end{table}

\subsection{Pre-training and fine-tuning pipeline}

At 10$\times$ magnification, we set the region size to be 4480$\times$4480 pixels in our experiments. Each region contains 400 non-overlapping 224$\times$224-pixel patches. This region size is selected to balance computational feasibility and the ability to identify high-level morphological structures. The computational cost of the transformer model increases quadratically with the input sequence length, hence, a larger region would be computationally unfeasible, while a smaller region may not be capable of capturing significant histological information. While choosing a region from WSI, we require at least 25\% of its patches to be foreground patches. Regions that are primarily background may only provide limited information, while selecting only regions that do not overlap with the background would greatly reduce the number of regions available for sampling. The extracted features of the foreground patches in a region and the zero paddings for the background patches as well as their corresponding locations are sent to the transformer model.

We follow the implementation of ViT \cite{Dosovitskiy2020} and set $L=12$ and $H=8$, as the number of layers and attention heads, respectively, in our experiments. This model is similar to ViT-Base but with fewer heads (8 versus 12) to accommodate the output dimension of the ResNet model ($d=512$).

40\% of patches from each region are randomly masked for restoration while pre-training. The learning rate is $4\times10^{-5}$, and is warmed up with the first 8,000 steps and then annealed according to the cosine schedule. For pre-training, 80\% of the data are for optimizing the transformer model, while the remaining 20\% are used to monitor the loss. With a batch size of 64, the model was pre-trained for a total of 400,000 iterations using the AdamW optimizer \cite{Loshchilov2019}.

For the downstream prediction tasks, more regions are sampled from a WSI during each iteration to ensure comprehensive coverage. The maximum overlap between any two regions is set to 50\% to avoid extensive repeated sampling. The learning rate for the baseline methods is $1\times10^{-3}$ for the survival prediction tasks and $3\times10^{-3}$ for the classification tasks. While fine-tuning the MaskHIT model, we used the same learning rate (i.e., $1\times10^{-3}$ or $3\times10^{-3}$) for the last fully connected layer as the baseline methods and a small learning rate (i.e., $1\times10^{-5}$) for the transformer backend. To prevent overfitting, an early stopping approach is utilized, with the training process ending after five stagnant epochs. 

\subsection{Baselines and evaluation}

Our downstream predictions include two different types of tasks: cancer survival prediction and cancer classification. The survival outcome is the time (in years) from cancer diagnosis to the recorded death. We used the negative log Cox Partial Likelihood Loss \cite{Cox1972}, which is defined as 
\begin{equation}
l(\theta)= -\frac{1}{N_{\delta=1}}\Sigma_{i:\delta_i=1}{\left(\hat{h_\theta}(x_i)-log\Sigma_{j\in R(T_i)}e^{\hat{h_\theta}(x_j)}\right)}
\label{eq}\end{equation}
where $\hat{h_\theta} (x_i )$  is the output (i.e., risk score) of a model with parameters $\theta$ and input $x_i$, $\delta=1$ means having the event (death), and $R(T_i )$ is patient $i$'s risk set (i.e., having not observed the event at time $T_i$). The Cross-entropy loss was used in the classification tasks and the model performance was evaluated using the macro-averaged AUC score.

We implemented two series of downstream evaluations. In the first series, we compared MaskHIT to several state-of-the-art (SOTA) MIL approaches. The first baseline method uses average pooling to aggregate feature vectors from patches (referred to as MIL-AP) \cite{Wulczyn2020,Jiang2021}. We also included three attention-based MIL methods, namely MIL-Attn \cite{Ilse2018}, DeepAttnMISL \cite{Yao2020}, and MHAttnSurv \cite{Jiang2023}. We kept the data sampling approaches exactly the same across these different methods, to demonstrate the effectiveness of MaskHIT for feature aggregation. 

In the first series of evaluations, we used 5-fold cross-validation for model evaluation. During each fold, 20\% of the data was set aside for testing purposes, 20\% was used for performance monitoring, and the remaining 60\% was utilized for model training. Specifically, for the survival prediction tasks, given the limited number of events, we noted a substantial variation in survival prediction performance depending on model initialization. As a result, we repeated the 5-fold cross-validation process 5 times and presented the average test c-index for a more stable evaluation.

In the second series, we compared MaskHIT to two other transformer-based approaches, H2T \cite{Vu2023} and HIPT \cite{Chen2022hipt}. The HIPT study is built upon transformer modules and pre-trained with large datasets. The H2T approach is a lightweight representation framework utilizing the co-localization of histological patterns. Given the different data preprocessing pipelines used in the HIPT study, we used the HIPT results directly from the corresponding publication. In addition to MaskHIT-10$\times$, we included MaskHIT-20$\times$ in this comparison as both baseline models were evaluated at 20$\times$ magnification. We also used the same data split as released in the HIPT repository for a more consistent comparison.

\section{Results}
\subsection{Visualization of self-attention}
\begin{figure}
    \centering
    \includesvg[width = 10cm]{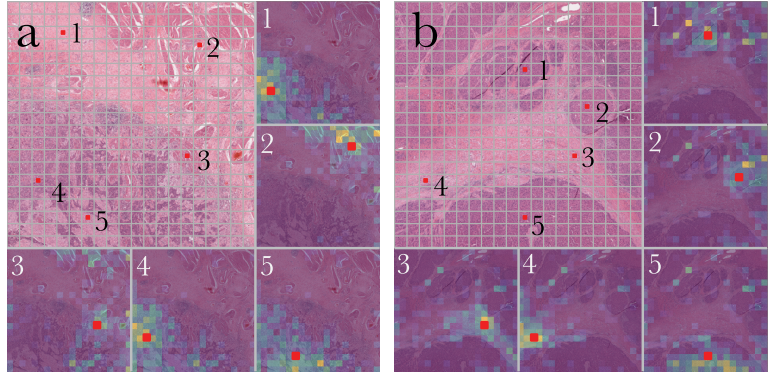}
    \caption{Attention map visualization for pre-trained model. a) BLCA. b) LIHC. The intensity of the color indicates the attention score with brighter colors representing a higher score and darker colors representing a lower score. Patches with a red dot are the query patches.}\label{fig:self-attn}
\end{figure}

After pre-training, we visualized the attention map learned by the MaskHIT model to evaluate if it could learn the underlying structure of WSI (\autoref{fig:self-attn}). Specifically, we chose several patches from one region on WSI as queries and calculated attention weights from other patches (as keys). These weights serve as an indicator of the extent to which the query patch relies on information from other patches. Our observations indicate that each patch tends to draw information from similar patches and those located in a close proximity. These observations support our objective to train a transformer model that can learn spatial relationships between patches in histopathological images.

\begin{table*}[]
\caption{Survival Prediction Performance, average c-index$\pm$ standard deviations}
\centering
\label{table:res-surv}
\resizebox{\textwidth}{!}{
\begin{tabular}{lcccccc}
\toprule
               & BRCA        & COAD        & LGG         & LUAD        & OV          & Avg         \\
               \midrule
MIL-AP \cite{Wulczyn2020}        & 0.579±0.010 & 0.600±0.017 & 0.643±0.019 & 0.551±0.007 & 0.568±0.021 & 0.588±0.005 \\
MIL-Attn \cite{Ilse2018}       & 0.566±0.025 & 0.583±0.014 & 0.655±0.007 & 0.560±0.014 & 0.587±0.016 & 0.590±0.004 \\
DeepAttnMISL \cite{Yao2020}   & 0.575±0.012 & 0.580±0.011 & 0.634±0.018 & \textbf{0.561±0.014} & 0.596±0.016 & 0.589±0.006 \\
MHAttn \cite{Jiang2023}         & 0.579±0.020 & 0.592±0.025 & 0.657±0.007 & 0.554±0.012 & 0.588±0.009 & 0.594±0.008 \\
MaskHIT (ours) & \textbf{0.602±0.012} & \textbf{0.608±0.022} & \textbf{0.685±0.009} & \textbf{0.561±0.009} & \textbf{0.602±0.008} & \textbf{0.612±0.004} \\ 
\bottomrule
\end{tabular}}
\end{table*}

\begin{table}[]
\caption{Cancer Subtype Classification Performance, average AUC$\pm$ standard deviations}
\label{table:res-cls}
\centering
\begin{tabular}{lccc}
\toprule
             & DHMC-RCC    & TCGA-Brain  & TCGA-BRCA   \\ 
\midrule
MIL-AP \cite{Wulczyn2020}       & 0.909±0.014 & 0.919±0.032 & 0.689±0.048 \\
MIL-Attn \cite{Ilse2018}     & 0.932±0.022 & 0.925±0.026 & 0.693±0.050 \\
DeepAttnMISL \cite{Yao2020} & 0.911±0.041 & 0.928±0.031 & 0.667±0.034 \\
MHATTN \cite{Jiang2023}       & 0.938±0.028 & 0.946±0.027 & 0.716±0.042 \\
MaskHIT (Ours)      & \textbf{0.958±0.017} & \textbf{0.964±0.013} & \textbf{0.735±0.026} \\ 
\bottomrule
\end{tabular}
\end{table}

\subsection{Comparison with MIL-based approaches}
Survival-1 results are summarized in \autoref{table:res-surv}. All experiments were conducted at 10$\times$ magnification. Results show that MaskHIT consistently outperforms or matches the baseline methods across different cancer types in terms of average cross-validation c-index. On average across the five cancer types, MaskHIT achieves a c-index of 0.612 with a standard deviation of 0.004 across repeated cross-validation, which is 3.0\% higher than the best-performing baseline MHAttn (0.594).

Classification-1 results are summarized in \autoref{table:res-cls}, which shows MaskHIT outperforms all the baseline approaches on all three classification tasks by a significant margin. MaskHIT achieves an AUC of 0.958, 0.964, and 0.735, for DHMC-RCC, TCGA-Brain and TCGA-BRCA, respectively. The average AUC improvement of MaskHIT over the best-performing baseline is 2.1\%, 1.9\%, and 2.7\%. 

\subsection{Comparison with transformer-based approaches}
\autoref{table:comp-surv} summarizes the comparison of the MaskHIT approach to the HIPT method in survival prediction tasks. H2T was excluded from this experiment as it was not evaluated for survival tasks in the original study. We find that MaskHIT-10$\times$ and MaskHIT-20$\times$ outperform HIPT on all cancer types evaluated in terms of average validation AUC with an average increase in c-index of 0.046 and 0.071, respectively.

\autoref{table:comp-cls} summarizes MaskHIT's performance compared to H2T and HIPT on breast cancer, kidney cancer, and lung cancer classification tasks using 25\% and 100\% of available data. MaskHIT-10$\times$ outperforms H2T and HIPT on breast and kidney tasks with 100\% training, but lags behind HIPT for lung cancer, while MaskHIT-20$\times$ consistently outperforms H2T and HIPT for all three cancer types with 100\% training. For kidney cancer with 25\% data, MaskHIT-20$\times$ performs similarly to HIPT (AUC of 0.973 versus 0.974). On average, MaskHIT-20$\times$ achieves an 8.5\% and 1.7\% increase in AUC compared to H2T and HIPT, respectively, with 25\% training. When trained with all available data, MaskHIT achieves an average increase of 4.9\% and 2.7\%, respectively.

\begin{table*}[]
\caption{Survival Prediction Performance Compared with Other Transformer-based Approaches. * Results from the original study.}
\label{table:comp-surv}
\centering
\resizebox{\textwidth}{!}{
\begin{tabular}{lcccccc}
\toprule
             & IDC         & CRC         & CCRCC       & PRCC        & LUAD        & STAD        \\ \midrule
HIPT\cite{Chen2022hipt}* & 0.634±0.050 & 0.608±0.088 & 0.642±0.028 & 0.670±0.065 & 0.538±0.044 & 0.570±0.081 \\
MaskHIT-10$\times$  & 0.642±0.029 & 0.684±0.083 & 0.649±0.049 & 0.743±0.081 & 0.599±0.027 & 0.621±0.045 \\
MaskHIT-20$\times$  & \textbf{0.645±0.031} & \textbf{0.697±0.101} & \textbf{0.665±0.072} & \textbf{0.823±0.082} & \textbf{0.615±0.027} & \textbf{0.640±0.027} \\ 
\bottomrule
\end{tabular}}
\end{table*}

\begin{table}[]
\caption{Cancer Subtype Classification Performance Compared with Other Transformer-based Approaches. * Results from the original study.}
\label{table:comp-cls}
\centering
\resizebox{\textwidth}{!}{
\begin{tabular}{lcccccc}
\toprule
\multirow{2}{*}{Architecture} & \multicolumn{2}{c}{Breast}     & \multicolumn{2}{c}{Kidney}     & \multicolumn{2}{c}{Lung}       \\ \cmidrule{2-7} 
                              & 25\% Training & 100\% Training & 25\% Training & 100\% Training & 25\% Training & 100\% Training \\ \midrule
H2T\cite{Vu2023}                   & 0.750±0.038   & 0.860±0.060    & 0.956±0.014   & 0.979±0.009    & 0.843±0.035   & 0.910±0.030    \\
HIPT\cite{Chen2022hipt}*                 & 0.821±0.069   & 0.874±0.060    & \textbf{0.974±0.012}   & 0.980±0.013    & 0.923±0.020   & 0.952±0.021    \\
MaskHIT-10$\times$                   & 0.867±0.042   & 0.931±0.047    & 0.972±0.012   & 0.986±0.008    & 0.894±0.030   & 0.945±0.026    \\
MaskHIT-20$\times$                   & \textbf{0.863±0.037}   & \textbf{0.932±0.045}    & 0.973±0.012   & \textbf{0.990±0.024}    & \textbf{0.929±0.026}   & \textbf{0.961±0.006}    \\ 
\bottomrule
\end{tabular}}
\end{table}

\subsection{Ablation studies}
We conducted additional investigations to determine the effects of pretext training, restoration loss function choice, number of sampled regions, and region size on downstream tasks. 
\subsubsection{Effect of pre-training}
\begin{figure}
    \centering
    \includesvg[width = 12cm]{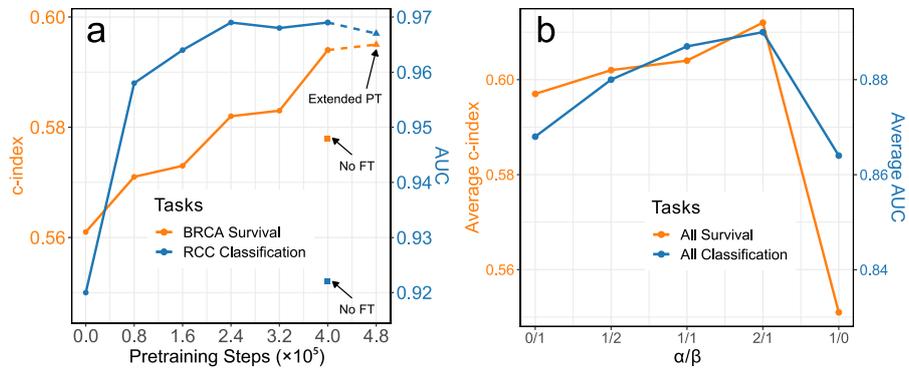}
    \caption{Effect of Pre-training and Choice of Loss Function Components on Downstream Tasks. a) Effect of pre-training. NO FT: No Fine Tuning - freeze transformer part when fine-tuning. Extended PT: extended Pre-Training - pre-train the final model for additional steps and then fine-tune. b) Effect of the restoration loss composition. $\alpha$: the coefficient of the L2 distance component; $\beta$: the coefficient of the InfoNCE loss component.}
    \label{fig:abl-pre-training}
\end{figure}

Using TCGA-BRCA as an example of the survival task, and DHMC-RCC as an example of the classification task, we show the effect of pre-training on downstream task performance (\autoref{fig:abl-pre-training}a). When the transformer model has not undergone any pre-training, it performs poorly on slide-level tasks; in fact, its performance can be worse than the baseline methods. Specifically, MIL-AP has a c-index of 0.579 in TCGA-BRCA survival prediction, while the raw transformer model only has a c-index of 0.561. This supports our intuition that using a transformer model without pre-training, given the number of parameters and the limited amount of labeled data, does not necessarily result in improved performance. 

Our findings demonstrate that with a moderate level of pretext training, transformer model performance improves. Our downstream tasks are performed after the vision transformer model has been pre-trained for a total of $4\times10^5$ steps. However, if we further pre-train this model for another 80 thousand steps (as indicated by extended PT, short for extended Pre-Training in \autoref{fig:abl-pre-training}a), it does not lead to any additional improvements in slide-level performance, suggesting that benefits of pre-training plateau after a sufficient amount of pre-training. Additionally, \autoref{fig:abl-pre-training}a reveals that if the transformer component is frozen for the downstream tasks (as indicated by No FT, short for No Fine-Tuning on Figure 3a), there is a significant decline in performance, which underscores the importance of using a small learning rate for the transformer model while fine-tuning for slide-level predictions.

\subsubsection{Effect of restoration loss function} \label{sssec:alpha-beta}
As our restoration loss consists of two components, L2 distance with weight $\alpha$ and InfoNCE loss with weight $\beta$, we evaluated how different combinations of these components could affect the downstream task performance. We reported the average c-index for the 5 survival prediction tasks and the average AUC for the 3 cancer subtype classification tasks in series 1, with the results summarized in \autoref{fig:abl-pre-training}b.

\autoref{fig:abl-pre-training}b indicates that increasing the weight of L2 distance leads to an improvement in both c-index and AUC, with the highest values achieved at $\alpha$=2 and $\beta$=1. However, when only L2 distance is utilized ($\alpha$=1, $\beta$=0), the model performance plunges on both tasks, suggesting that L2 distance alone is inadequate for computing the restoration loss and that the inclusion of the contrastive component is necessary to learn contextual information from the histology images.

\subsubsection{Effect of the number of patches and the number of regions}

\begin{figure}
    \centering
    \includesvg[width = 12cm]{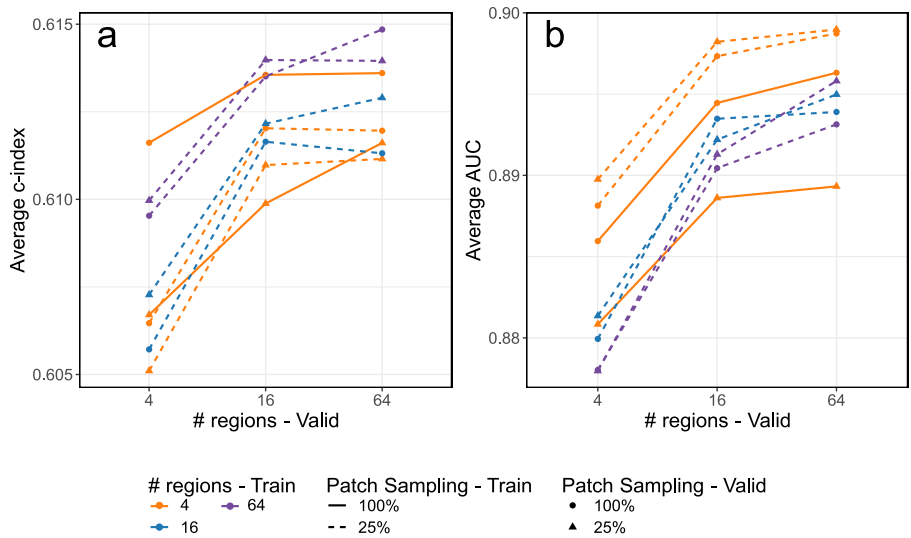}
    \caption{Effect of Region and Patch Sampling at Training and Validation (valid) Time on Prediction Performance. a) Average c-index}
    \label{fig:aba-number}
\end{figure}

We further evaluated the effects of region and patch sampling procedures on the downstream tasks. The results are presented in \autoref{fig:aba-number}a for survival tasks and \autoref{fig:aba-number}b for classification tasks. The solid orange lines represent models trained with 4 regions sampled from each WSI and 100\% of patches sampled from each region. We will refer to this sampling schema as 4 regions with 100\% coverage, or 4$\times$100\% for short. The series with round dots are models evaluated with 100\% coverage, and the series with triangles are models evaluated with 25\% coverage. We found that when more regions are sampled from each WSI during the evaluation time, the average test performance also increases. Additionally, sampling with 100\% coverage at evaluation time outperforms 25\% coverage by a large margin.

Focusing on the dashed lines, which refer to models trained with 25\% coverage, we can evaluate the effect of the number of regions sampled at training time on the whole slide-level prediction performance. For the survival prediction task, we found that sampling more regions from each WSI is associated with an increased test c-index. However, the trend is the opposite for the classification tasks, as a higher number of regions sampled at training time is associated with worse test AUC. The effect of patch sampling at evaluation time is inconsistent across our experiments, but the number of regions sampled at evaluation time is mostly associated with better performance.

Among all these training/evaluation scenarios, the best test performance for the survival prediction tasks is 0.615, achieved when training the models using 64$\times$25\%, and evaluating with 64$\times$100\%. The best-performing model for the classification task is trained with 4$\times$25\% and evaluated with 64$\times$25\%, with an average AUC of 0.899.

\subsubsection{Effect of region size}

 \begin{figure}
    \centering
    \includesvg[width = 12cm]{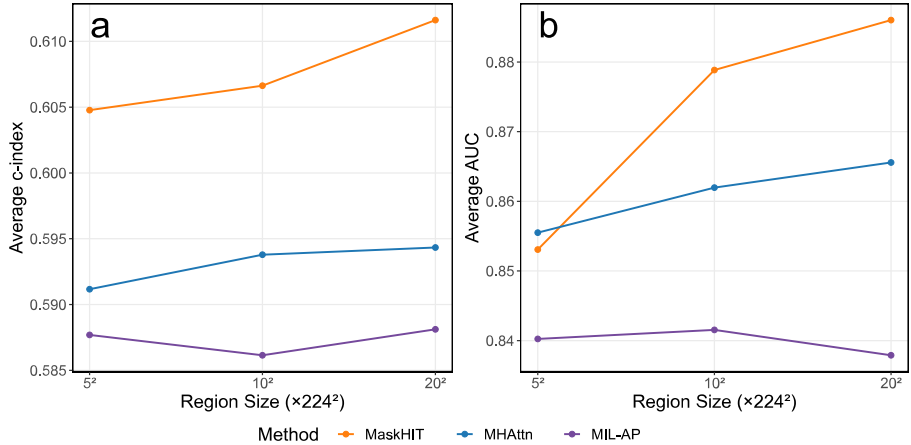}
    \caption{Effect of Region Size on MaskHIT, MHAttn and MIL-AP on Prediction Performance. a) Average c-index of survival prediction tasks. b) Average AUC of classification tasks.}
    \label{fig:aba-size}
\end{figure}

We evaluated the effect of region size on the prediction performance for survival tasks and classification tasks, and presented the results in \autoref{fig:aba-size}a and Figure \autoref{fig:aba-size}b, respectively. In addition to MaskHIT, we included two additional methods as baselines: the plain MIL method, MIL-AP, and the attention-based method, MHAttn. We experimented with three region sizes: the small region with $5^2\times224^2$ pixels, the medium region with $10^2\times224^2$ pixels, and the large region with $20^2\times224^2$ pixels. We trained the models using 4 large regions, and evaluated them with either 64 small regions, 16 medium regions, or 4 large regions, to maintain consistency in the total number of patches across experiments. 

The results indicate that, for MaskHIT, larger region sizes result in better survival and classification performance, with the trend being more evident for the classification tasks. For the MHAttn method, an increase in performance is also observed when evaluating using larger regions, but the trend is mild compared to the MaskHIT model. The results for the MIL-AP method are mixed without clear trend. This ablation study demonstrates that the MaskHIT model benefits from larger region sizes compared to MIL-based approaches, and a region size of 4480$\times$4480 pixels at 10$\times$ resolution is a good candidate. Although the performance does not saturate with this region size, increasing it further would quadratically increase the computational costs and was not explored.

\subsection{Model visualization}

 \begin{figure}
    \centerline{\includegraphics[width=\columnwidth]{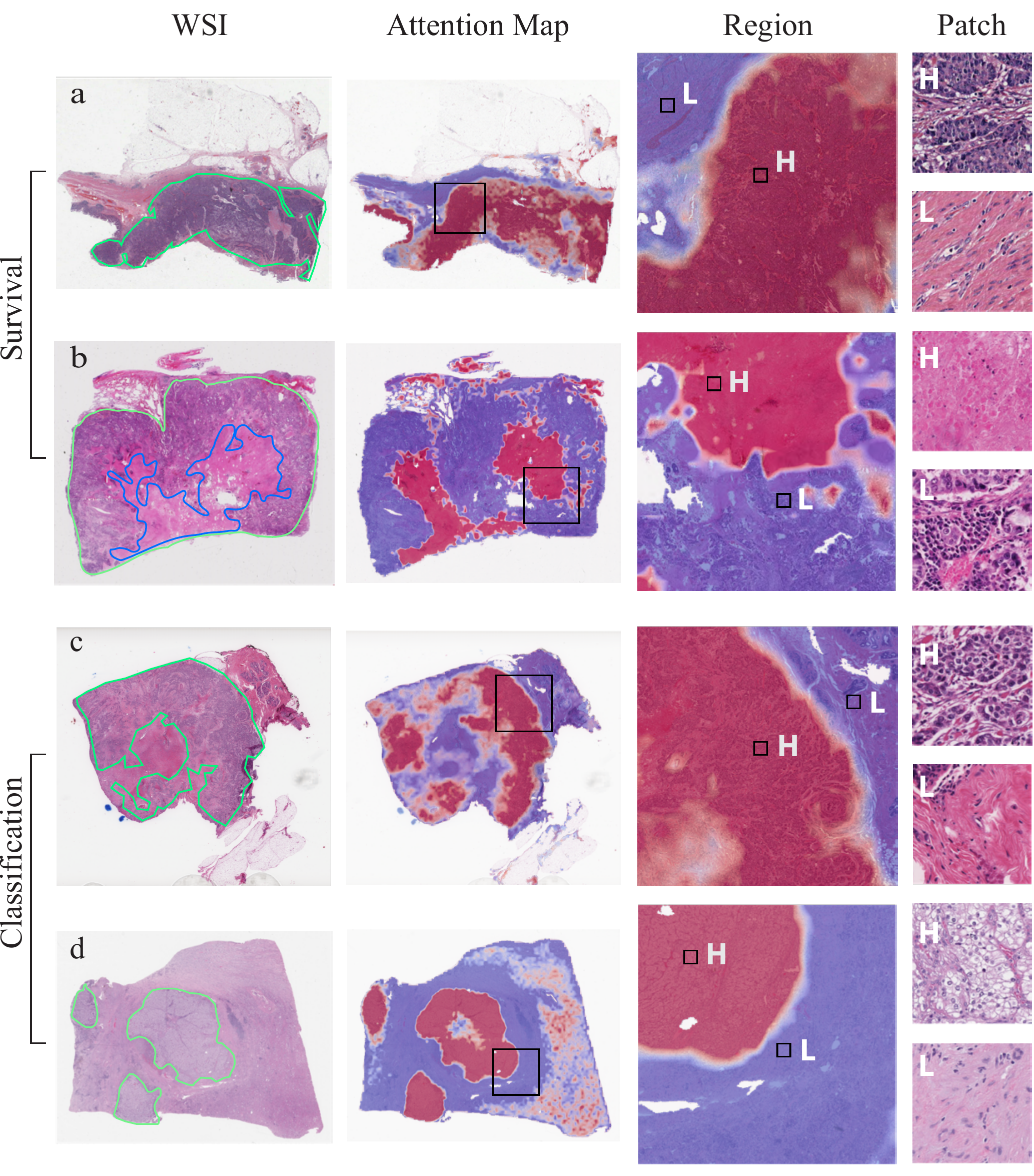}}
    \caption{Visualization of attention maps. Region size: 4480\textmu m, Patch size: 224\textmu m. a) COAD; b) LUAD; c) BRCA; d) KIRC. Red color for higher attention (H), blue color for lower attention (L). First column labeled with ``WSI" shows pathologist's annotations of tumor in green color and necrosis in blue color.}
    \label{fig:vis}
\end{figure}

To understand MaskHIT's predictions, an analysis of the attention maps from selected WSIs was conducted. The procedure involved averaging out attention weights obtained from different attention heads for each layer, followed by applying recursive multiplication of this weight matrix across all attention layers to calculate overall attention weights. We extracted the attention weights of the class token, which served as a representation of the information incorporated in the class token from each patch. 

We calculated attention maps using both pre-trained model and model fine-tuned for the downstream tasks. Their difference highlights the shift in the class token's attention from the targetless pre-training task to the guided WSI-level prediction tasks, thus providing insight into the model's attention mechanism during the prediction process.

\autoref{fig:vis} presents a sample from each of the COAD, LUAD, BRCA, and KIRC datasets from TCGA, with attention maps obtained using the MaskHIT-10$\times$ model for survival tasks in COAD and LUAD, and molecular subtype and cancer classification tasks in BRCA and KIRC, respectively. The COAD example shows good alignment between the higher attention areas detected by the model and the tumor region identified by the pathologist. In the LUAD example, where the tumor extends to the entire slide, the model pays more attention to regions featuring tumor necrosis, which is typically associated with favorable patient outcomes. Additionally, in the BRCA and KIRC examples, the model's higher attention areas align well with pathologist annotations of the tumor region, demonstrating the model's ability to successfully relocate attention for each specific downstream task.

\section{Discussion and Conclusion}
Our study proposes a novel approach for whole slide image representation using a transformer model and multiple instance learning. Our unsupervised training pipeline for the transformer model, which restores randomly masked patches, outperforms existing methods for both survival prediction and classification tasks. Visualization of the self-attention maps demonstrates that our pre-training task enables the transformer model to learn spatially and morphologically relevant associations among patches. The adaption attention map after fine-tuning the transformer model for downstream tasks demonstrates its ability to recognize clinically meaningful visual patterns. Our ablation results support the positive effects of pretext training and show that our transformer-based approach benefits from using larger region sizes to capture positional information of patches.

We evaluated MaskHIT on various types of tasks, including survival prediction, subtype classification, grade classification, and molecular subtype classification. Subtype and grade classification are usually easier tasks as they are typically based on visual patterns, such as cell size and shape, mitotic rate, and the presence or absence of certain cell types. However, histopathological examination alone is not definitive for determining the molecular subtype of the tumor, which generally requires molecular or genomic analysis of the tumor tissue. Moreover, survival prediction using histological images is challenging given the lack of definitive prognostic features for most cancer types. Despite these variations, we found that MaskHIT consistently outperforms existing methods, which demonstrates the robustness of our approach for a wide spectrum of tasks.

One limitation of our approach is that we still rely on the MIL framework to represent WSIs. As demonstrated in Figure 6, the vision transformer model can benefit from a larger region size. Although the transformer model enables us to represent a large region with 20 million pixels (or 80 million pixels at 20$\times$ resolution) from WSIs, further expanding the region size to the entire slide is not computationally feasible at this stage. Therefore, although we can learn the spatial arrangements of patches within a relatively large region, we may still lose information on the higher region-level positions. While the HIPT model was designed to solve this problem through the hierarchical use of transformer modules, pre-training the transformer model beyond the region level is still a challenging task given the limited number of slides available for pre-training. Future studies may focus on developing novel pretext tasks or expanding the collection of histopathological images to overcome this challenge.

Moreover, the proposed MaskHIT model can be further enhanced through utilizing better backbone models specialized for feature extraction from medical images for patch feature representation. We anticipate that the proposed MaskHIT model would be effective for various tasks and applications in computational pathology, which we plan to explore in future work.

\textbf{\bibliographystyle{ieeetr}
\bibliography{reference}}

\end{document}